\documentclass{sig-alternate}

\usepackage{array}
\usepackage{mdwmath}
\usepackage{mdwtab}
\usepackage{url}
\usepackage{bm}
\usepackage{rotating}

\usepackage{amsmath,amssymb}
\newcommand{\st}{\mathrm{s.t.}}
\newcommand{\diag}{\mathrm{diag}}
\newcommand{\tr}{\mathrm{tr}}
\newtheorem{theorem}{Theorem}

\newcommand{\tabincell}[2]{\begin{tabular}{@{}#1@{}}#2\end{tabular}}
\usepackage{booktabs}
\usepackage{multirow}
\usepackage{algorithm}
\usepackage{algorithmic}

\usepackage{graphicx,subfig}
\usepackage{float}

\begin{document}
\title{Unsupervised Feature Selection with\\ Adaptive Structure Learning}
\numberofauthors{2}
\author{
\alignauthor Liang Du\\
\affaddr{ State Key Laboratory of Computer Science, \\
Institute of Software, Chinese Academy of Sciences}\\
\email{duliang@ios.ac.cn}
\alignauthor Yi-Dong Shen\\
\affaddr{ State Key Laboratory of Computer Science, \\
Institute of Software, Chinese Academy of Sciences}\\
\email{ydshen@ios.ac.cn}
}

\maketitle

\abstract{
The problem of feature selection has raised considerable interests in the past decade. Traditional unsupervised methods select the features which can faithfully preserve the intrinsic structures of data, where the intrinsic structures are estimated using all the input features of data. However, the estimated intrinsic structures are unreliable/inaccurate when the redundant and noisy features are not removed. Therefore, we face a dilemma here: one need the true structures of data to identify the informative features, and one need the informative features to accurately estimate the true structures of data. To address this, we propose a unified learning framework which performs structure learning and feature selection simultaneously. The structures are adaptively learned from the results of feature selection, and the informative features are reselected to preserve the refined structures of data. By leveraging the interactions between these two essential tasks, we are able to capture accurate structures and select more informative features.  Experimental results on many benchmark data sets demonstrate that the proposed method outperforms many state of the art unsupervised feature selection methods.}

\section{Introduction}
\begin{figure*}[ht]
\centering
\includegraphics[width=0.85\textwidth]{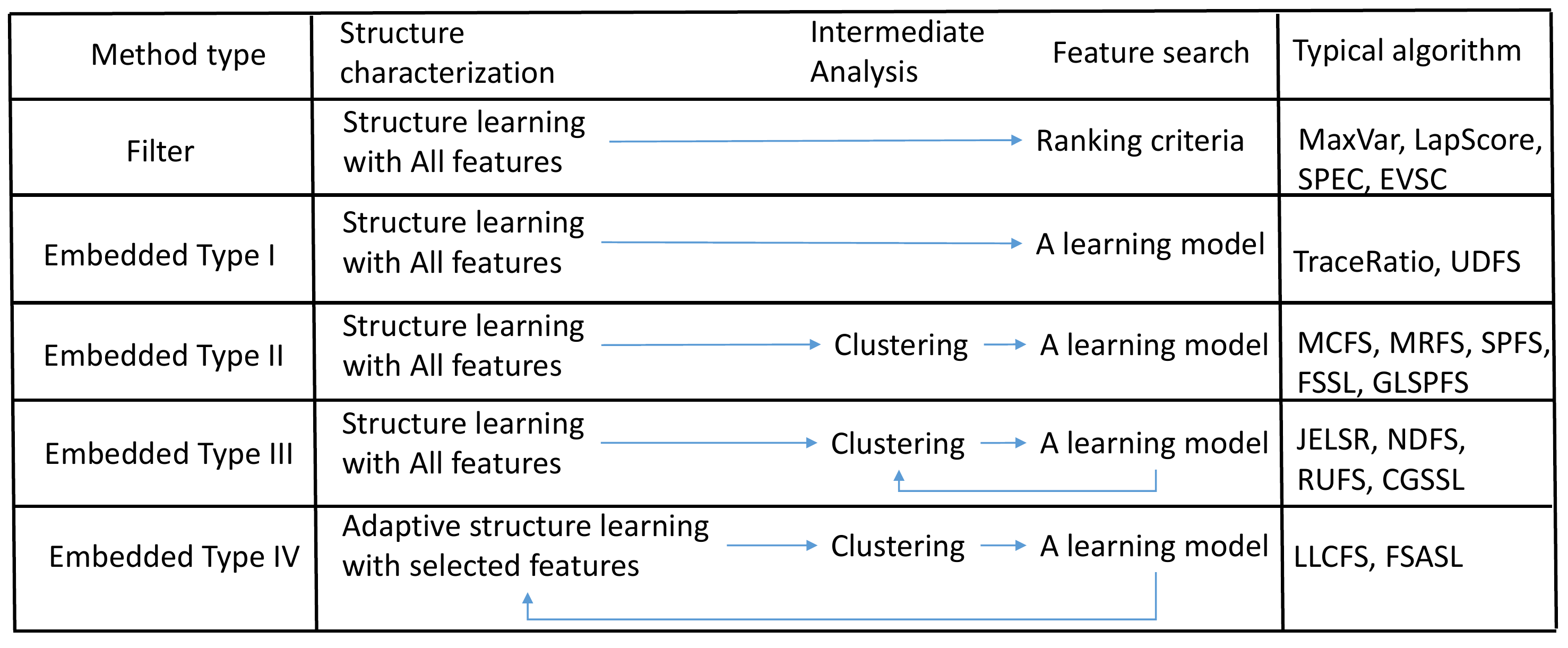}
\caption{An illustration of unsupervised filter methods and four type embedded methods.}
\label{ufs_tree}
\end{figure*}
Real world applications usually involve big data with high dimensionality, presenting great challenges such as the curse of dimensionality, huge computation and storage cost. To tackle these difficulties, feature selection techniques are developed to keep a few relevant and informative features. 
According to the availability of label information, these algorithms can be categorized into supervised \cite{lasso}, \cite{robnik2003theoretical}, \cite{mifs}, \cite{rfs_l21}, semi-supervised \cite{zhao2007semi}, \cite{xu2010discriminative} and unsupervised algorithms \cite{dy2004feature}, \cite{mcfs}. Compared to supervised or semi-supervised counterparts, unsupervised feature selection is generally more challenging due to the lack of supervised information to guild the search of relevant features.

Unsupervised feature selection has attracted much attention in recent years
and a number of algorithms have been proposed \cite{lapscore,mcfs,zhao2013similarity,udfs,glspfs}. Without class label, unsupervised feature selection chooses features that can effectively reveal or maintain the underlying structure of data. Recent research on feature selection and dimension reduction has witnessed that several important structures should be preserved by the selected features. These important structures include, but not limited to, the global structure \cite{zhao2013similarity,glspfs}, the local manifold structure \cite{lapaofs,jelsr2} and the discriminative information \cite{udfs,ndfs}. And these structures can be captured by widely used models in the form of graph, such as, the sample pairwise similarity graph \cite{zhao2013similarity}, the $k$-nn graph \cite{lapscore}, the global integration of local discriminant model \cite{udfs,yang2013discriminative}, the local linear embedding (LLE) \cite{glspfs}.

Clearly, many of existing unsupervised feature selection methods rely on the structure characterization through some kind of graph, which can be computed within the \emph{original} feature space. And once the graph is determined, it is \emph{fixed} in the next procedures, such as sparse spectral regression \cite{sr}, to guide the search of informative features. Consequently, the performance of feature selection is largely determined by the effectiveness of graph construction. Ideally, such graphs should be constructed only using the informative feature subset rather than all candidate features. Unfortunately, the desired subset of features is unknown in advance, and the irrelevant or noisy features would be inevitably introduced in many real applications. As a result, unrelated or noisy features will have an adverse effect on the characterization of the structures and henceforth hurt the following feature selection performance. 

In unsupervised scenario, this is actually the chicken-and-egg problem between \emph{structure characterization} and \emph{feature selection}. Facing with such dilemma, we propose to perform feature selection and structure learning in a unified framework, where each sub task can be iteratively boosted by using the result of the other one. Concretely, the global structure of data is captured within the sparse representation framework, where the reconstruction coefficient is learned from the selected features. The local manifold structure is revealed by a probabilistic neighborhood graph, where the pairwise relationship is also determined by the selected features. When the global and local structures are given in the form of graph Laplacians, we seek the relevant features via sparse spectral regression with the help of graph embedding for cluster analysis. In this way, both the global and local structure of data can be better captured by only using the selected features; Moreover, with the refined characterization of the structure, a better search of the informative features could also be expected.

It is worthwhile to highlight several aspects of the proposed approach here
\begin{enumerate}
\item Based on the different learning paradigms for unsupervised feature selection, we investigate most of existing unsupervised embedded methods and further classify them into four closely related but different types. These analyses provide more insight into what should be further emphasized on the development of more essential unsupervised feature selection algorithm.
\item We propose a novel unified learning framework, called unsupervised Feature Selection with Adaptive Structure Learning (FSASL in short), to fulfil the gap between two essential sub tasks, i.e. structure learning and feature learning. In this way, these two tasks can be mutually improved.
\item  Comprehensive experiments on benchmark data sets show that our method achieves statistically significant improvement over state-of-the-art feature selection methods, suggesting the effectiveness of the proposed method.
\end{enumerate}

The rest of this paper is arranged as follows. We review the related work in Section 2. Then we present our proposed formulation and the developed solution in Section 3. Discussions are introduced in Section 4. Extensive experiments are conducted and analyzed in Section 5. Section 6 concludes this paper with future work.

\section{Related Works}
In this section, we mainly review most existing unsupervised feature selection methods, i.e. filter and embedded methods. Unsupervised filter methods pick the features one by one based on certain evaluation criteria, where no learning algorithm is involved. The typical methods include: max variance (MaxVar) \cite{krzanowski1987selection}, Laplacian score (LapScore) \cite{lapscore}, spectral feature selection (SPEC) \cite{spec}, feature selection via eigenvalue sensitive criterion (EVSC) \cite{mcfs}. A common limitation of these approaches is the correlation among features is neglected \cite{alelyani2013feature}.

Unsupervised embedded approaches are developed to perform feature selection and fit a learning model simultaneously. Based on the different sub-steps involved in the feature selection procedure, these embedded methods can be further divided into four different types as illustrated in Figure \ref{ufs_tree}.

The first type of embedded methods first detect the structure of the data and then directly select those features which is used to best preserve the enclosed structure. The typical methods include: trace ratio (TraceRatio) \cite{traceratio} and unsupervised discriminative feature selection (UDFS) \cite{udfs}. TraceRatio is prone to select redundant features \cite{glspfs} and the learning model of UDFS is often too restrictive \cite{qian2013robust}.

The second type of embedded methods first construct various graph Laplacians to capture the data structure, then flat the cluster structure via graph embedding, and finally use the sparse spectral regression \cite{sr} to select those features that are best aligned to the embedding. Instead of directly selecting features as the first type, these approaches resorted to an \emph{intermediate cluster analysis sub-step} to reveal the cluster structure for guiding the selection of features. The cluster structure discovered by either the graph spectral embedding or other clustering module can be seen as an approximation of the unseen labels. The typical methods include: multi-cluster feature selection (MCFS) \cite{mcfs}, minimum redundancy spectral feature selection (MRSF) \cite{mrsf},  similarity preserving feature selection (SPFS) \cite{zhao2013similarity}, and joint feature selection and subspace learning (FSSL
) \cite{fssl}, global and local structure preserving feature selection (GLSPFS) \cite{glspfs}. 

Unlike the second type methods, the clustering analysis in the third type of embedded methods is co-determined by the embedding of the graph Laplacian and the \emph{adaptive discriminative regularization} \cite{yang2011nonnegative}, \cite{yang2013discriminative}, which can be obtained from the result of sparse spectral regression. By using the feedback from feature selection, the whole learning procedure can provide better cluster analysis, and vice versa. The typical methods include: joint embedding learning and spectral regression (JELSR) \cite{jelsr}, \cite{jelsr2}, nonnegative discriminative feature selection (NDFS) \cite{ndfs}, robust unsupervised feature selection (RUFS) \cite{qian2013robust}, feature selection via clustering-guided sparse structural learning (CGSSL) \cite{cgssl}.

The fourth type of embedded methods try to feed the result of feature selection into the structure learning procedure for improving the quality of structure learning. In \cite{llcfs}, a feature selection method is proposed for local learning-based clustering (LLCFS), which incorporates the relevance of each feature into the built-in regularization of the local learning model, where the induced graph Laplacian can be iteratively updated. However, LLCFS uses the discrete $k$-nearest neighbor graph and does not optimize the same objective function in structure learning and feature search.

It can be seen that all these above methods (except LLCFS) share a common drawback: they use all features to estimate the underlying structures of data. Since the redundant and noisy features are unavoidable in real world applications, that is also why we need feature selection, the learned structures using all features will also be contaminated, which would degrade the performance of feature selection. By leveraging the coherent interactions between structure learning and feature selection, our proposed method FSASL seamlessly integrates them into a unified framework, where the result of one task is used to improve the other one. 

\section{Unsupervised Feature Selection with Adaptive Structure Learning}
Let $\mathbf{X} = \{\mathbf{x}_1 ,..., \mathbf{x}_n\} \in \mathcal{R}^{d \times n}$ denotes the data matrix, whose columns correspond to data instances and rows to features. The generic problem of unsupervised feature selection is to find the most informative features. With the absence of class label to guild the search of relevant features, the data represented with the selected features should well preserve the intrinsic structure as the data represented by all the original features. 

To achieve this goal, we propose to jointly perform unsupervised feature selection and data structure learning simultaneously, where both the global and local structure are adaptively updated using the result of current feature selection.

We first summarize some notations used throughout this paper. We use bold uppercase characters to denote matrices, bold lowercase characters to denote vectors. For an arbitrary matrix $\mathbf{A} \in \mathcal{R}^{r \times t}$, $\mathbf{a}_i$ means the $i$-th column vector of $\mathbf{A}$ and $\mathbf{a}_j^T$ means the $j$-th row vector of $\mathbf{A}$, $\mathbf{A}_{ij}$ denotes the $(i,j)$-th entry of $\mathbf{A}$. The $\ell_{2,1}$-norm is defined as $||\mathbf{A}||_{21} = \sum_{i=1}^{r}\sqrt{\sum_{j=1}^{t} \mathbf{A}_{ij}^2}$.

\subsection{Adaptive Global Structure Learning}
Over the past decades, a large number of algorithms have been proposed based on the analysis of the global structure of data, such as the Principal Component Analysis (PCA) and the Maximum Variance (MaxVar). Recently, the global pairwise similarity (e.g., with a Gaussian kernel) between high-dimensional samples has demonstrated promising performance for unsupervised feature selection \cite{zhao2013similarity,glspfs}. However, such dense similarity becomes less discriminative for high dimension data, especially when there are many unfavorable features in the original high dimensional space.

Inspired by the recent development on compressed sensing and sparse representation \cite{wright2009robust}, we use the sparse reconstruction coefficients to extract the global structure of data. In sparse representation, each data sample $\mathbf{x}_i$ can be approximated as a linear combination of all the other samples, and the optimal sparse combination weight matrix $\mathbf{S} \in \mathcal{R}^{n\times n}$ can be obtained by solving the following problem
\begin{align}\label{sr}
\min_{\mathbf{S}} \quad \sum_{i=1}^{n} \left(||\mathbf{x}_i - \mathbf{X} \mathbf{s}_i||^2 + \alpha ||\mathbf{s}_i||_1\right) \quad \st \quad \mathbf{S}_{ii} = 0
\end{align}
where $\alpha$ is used to balancing the sparsity and the reconstruction error. Compared with the pairwise similarity, the sparse representation is naturally discriminative: among all the candidates samples, it selects the samples which most compactly expresses the target and rejects all other possible but less compact candidates \cite{wright2009robust}.

Clearly, the selected features should preserve such global and sparse reconstruction structure. To achieve this, we introduce a row sparse feature selection and transformation matrix $\mathbf{W} \in \mathcal{R}^{d \times c}$ to the reconstruction process, and get
\begin{align}\label{global_asl}
\min_{\mathbf{S}, \mathbf{W} } \quad & \sum_{i=1}^{n}||\mathbf{W}^T \mathbf{x}_i - \mathbf{W}^T \mathbf{X} \mathbf{s}_i||^2 + \alpha ||\mathbf{S}||_1 + \gamma ||\mathbf{W}||_{21}\\
\st  \quad & \mathbf{S}_{ii} = 0, \mathbf{W}^T \mathbf{X} \mathbf{X}^T \mathbf{W} = \mathbf{I} \nonumber
\end{align}
where $\gamma$ is regularization parameter. Compared with the Eq.\eqref{sr}, the benefits of Eq.\eqref{global_asl} are two folds: 1) The global structure captured by $\mathbf{S}$ can be used to guide the search of relevant features; 2) By largely eliminating the adverse effect of noisy and unfavorable features, the global structure can also be better estimated.

\subsection{Adaptive Local Structure Learning}
The importance of preserving local manifold structure has been well recognized in the recent development of unsupervised feature selection algorithms, especially considering that high-dimensional data often presents a low-dimensional manifold structure \cite{lapscore,mcfs,glspfs}. To detect the underlying local manifold structure, these algorithms usually first construct a $k$-nearest neighbor graph and then compute the graph Laplacian with different models. Clearly, both the $k$-nn graph and the graph Laplacian are determined by \emph{all the relevant and irrelevant features}. As a result, the manifold structure captured by such graph Laplacian would be inevitably affected by the redundant and noisy features. Moreover, the iterative updating of discrete neighborhood relationship using the result of feature selection still suffers from the lack of theoretical guarantee of its convergence \cite{llcfs,takeuchi2011target}.

Instead of using the graph Laplacian with the determinate neighborhood relationship, we introduce to directly learn a euclidean distance induced probabilistic neighborhood matrix. For each data sample $\mathbf{x}_i$, all the data points $\{\mathbf{x}_j\}_{j=1}^{n}$ are considered as the neighborhood of $\mathbf{x}_i$ with probability $\mathbf{P}_{ij}$, where $\mathbf{P} \in \mathcal{R}^{n \times n}$ can be determined by solving the following problem:
\begin{equation}
\min_{\mathbf{P}} \sum_{i,j}^{}(||\mathbf{x}_i - \mathbf{x}_j ||_2^2 \mathbf{P}_{ij} + \mu \mathbf{P}_{ij}^2),\textbf{ } \st \textbf{ } \mathbf{P}\mathbf{1}_n = \mathbf{1}_n, \mathbf{P} \geq 0
\end{equation}
where $\mu$ is the regularization parameter. The regularization term is used to 1) avoid the trivial solution; 2) add a prior of uniform distribution.
It can be found that a large distance $||\mathbf{x}_i - \mathbf{x}_j ||_2^2$ will lead to a small probability $\mathbf{P}_{ij}$. With such nice property, the estimated weight matrix $\mathbf{P}$ and the induced Laplacian $\mathbf{L}_{\mathbf{P}} = \mathbf{D}_{\mathbf{P}} - (\mathbf{P} + \mathbf{P}^T)/2$ can be used for local manifold characterization, where $\mathbf{D}_{\mathbf{P}}$ is a diagonal matrix whose $i$-th diagonal element is $\sum_{j} (\mathbf{P}_{ij} + \mathbf{P}_{ji})/2$.

To leverage the result of feature selection and iteratively improve the probabilistic neighborhood relationship, we also introduce the feature selection and transformation matrix $\mathbf{W}$ as used in global structure adaptive learning, and we get
\begin{align}\label{local_asl}
\min_{\mathbf{P}, \mathbf{W}} & \quad \sum_{i,j}^{n}(||\mathbf{W}^T\mathbf{x}_i - \mathbf{W}^T \mathbf{x}_j ||_2^2 \mathbf{P}_{ij} + \mu \mathbf{P}_{ij}^2) + \gamma ||\mathbf{W}||_{21} \\
\st & \quad \mathbf{P}\mathbf{1}_n = \mathbf{1}_n, \mathbf{P} \geq \mathbf{0}, \mathbf{W}^T \mathbf{X} \mathbf{X}^T \mathbf{W} = \mathbf{I} \nonumber
\end{align}
With the sparsity on $\mathbf{W}$, the irrelevant and noisy features can be largely removed, thus we can learn a better probabilistic neighborhood graph for local structure characterization based on the result of feature selection, i.e. $\mathbf{W}^T \mathbf{X}$. Moreover, we aim to seek those features to preserve the local structure encoded by $\mathbf{P}$. Thus, the optimization problem in Eq. \eqref{local_asl} can be used to perform feature selection and local structure learning, simultaneously. 
 
\subsection{Unsupervised Feature Selection with Adaptive Structure Learning}
Based on the two adaptive structure learning models presented in Eq. \eqref{global_asl} and Eq. \eqref{local_asl}, we propose a novel unsupervised feature selection method by solving the following optimization problem,
\begin{align}\label{opt_fsasl}
\min_{\mathbf{W}, \mathbf{S}, \mathbf{P}} \quad & \quad \left(||\mathbf{W}^T\mathbf{X} - \mathbf{W}^T\mathbf{X} \mathbf{S}||^2 + \alpha ||\mathbf{S}||_1 \right) \\
+& \quad \beta \sum_{i,j}^{n} \left(||\mathbf{W}^T \mathbf{x}_i - \mathbf{W}^T\mathbf{x}_j ||^2 \mathbf{P}_{ij} + \mu \mathbf{P}_{ij}^2 \right) + \gamma ||\mathbf{W}||_{21}  \nonumber  \\
\st \quad & \mathbf{S}_{ii} = 0, \mathbf{P}\mathbf{1}_n = \mathbf{1}_n, \mathbf{P} \geq \mathbf{0}, \mathbf{W}^T \mathbf{X} \mathbf{X}^T \mathbf{W} = \mathbf{I} \nonumber
\end{align}
where $\beta$ and $\gamma$ are regularization parameters balancing the fitting error of global and local structure learning in the first and second group and the sparsity of the feature selection matrix in the third group. 

It can be seen that when both $\mathbf{S}$ and $\mathbf{P}$ are given, our method selects those features to well respect both the global and local structure of data. When the feature selection matrix $\mathbf{W}$ is given, our method learns the global and local structure of data in a transformed space, i.e. $\mathbf{W}^T\mathbf{X}$, where the adverse effect of noisy features is largely alleviated with sparse regularization. In this way, these two essential tasks can be boosted by the other one within a unified learning framework. Since both the global and local structure can be adaptively refined according to the result of feature selection, we call Eq. \eqref{opt_fsasl} unsupervised Feature Selection with Adaptive Structure Learning (FSASL).

\subsection{Optimization Algorithm}
Because the optimization problem in Eq. \eqref{opt_fsasl} comprises three different variables with different regularizations and constraints, it is hard to derive its closed solution directly. Thus we derive an alternative iterative algorithm to solve the problem, which converts the problem with a couple of variables ($\mathbf{S}$, $\mathbf{P}$ and $\mathbf{W}^T$) into a series of sub problems where only one variable is involved.

First, when $\mathbf{W}$ and $\mathbf{P}$ are fixed, we need to solve $n$ decoupled sub problems in the following form:
\begin{align}\label{opt_s}
\min_{\mathbf{s}_i} \quad ||\mathbf{x}_i^{'} - \mathbf{X}^{'} \mathbf{s}_i||^2 + \alpha |\mathbf{s}_i|, \quad \st \quad \mathbf{S}_{ii} = 0
\end{align}
where $\mathbf{X}^{'}$ is the new transformed data by projecting the relevant features into a low dimension space, and $\mathbf{X}^{'}= \mathbf{W}^T \mathbf{X}$. The above LASSO problem can be efficiently solved by routine optimization tools, e.g. proximal methods \cite{bach2012optimization,liu_slep}.

Next, when $\mathbf{W}^T$ and $\mathbf{S}$ are fixed, we need to solve $n$ decoupled sub problems in the following form:
\begin{align}\label{sub_pp}
\min_{\mathbf{p}_i^T} \quad& \sum_{j=1}^{n} ||\mathbf{x}_i^{'} - \mathbf{x}_j^{'} ||^2 \mathbf{P}_{ij} + \mu ||\mathbf{P}_{ij}||^2, \\
\st \quad& \mathbf{1}_n^T \mathbf{p}_i = 1, \mathbf{P}_{ij} \geq 0 \nonumber
\end{align}
Denote $\mathbf{A} \in \mathcal{R}^{n \times n}$ be a square matrix with $\mathbf{A}_{ij} = -\frac{1}{2\mu} ||\mathbf{x}_i^{'} - \mathbf{x}_j^{'} ||^2$, then the above problem can be written as follows
\begin{align}\label{opt_proj}
\min_{\mathbf{p}_i^T} \quad \frac{1}{2}||\mathbf{p}_i^T - \mathbf{a}_i^T||^2, \quad \st \quad \mathbf{p}_i^T\mathbf{1}_n  = 1, 0 \leq \mathbf{p}_{ij}^T \leq 1
\end{align} 
where $\mathbf{p}_i^T$ is the $i$-th row of $\mathbf{P}$. The above euclidean projection of a vector onto the probability simplex can be efficiently solved by Algorithm \ref{alg_proj} without iterations. More details can be found in Eq. \eqref{optval_P}.
\begin{algorithm}
	\caption{The optimization algorithm of Eq. \eqref{opt_proj}}
	\label{alg_proj}
	\begin{algorithmic}		
	\REQUIRE{$\mathbf{a}$}
	\STATE{sort $\mathbf{a}$ into $\mathbf{b}$ where $b_1 \geq b_2 \geq, ..., b_n$}
	\STATE{find $\rho = \max \{ 1 \leq j \leq n: b_j + \frac{1}{j}(1 - \sum_{i=1}^{j} b_i) > 0 \}$}
	\STATE{define $z = \frac{1}{\rho}(1 - \sum_{i=1}^{\rho} b_i)$}
	\ENSURE{$\mathbf{p}$ with $p_j = \max\{a_j + z, 0\}, j = 1, ..., n$}
	\end{algorithmic}
\end{algorithm}

Next, when $\mathbf{S}$ and $\mathbf{P}$ are fixed, we need to solve the following problem:
\begin{align}
\min_{\mathbf{W}} \quad& ||\mathbf{W}^T\mathbf{X} - \mathbf{W}^T\mathbf{X} \mathbf{S}||^2 + \beta \sum_{i,j}^{n} ||\mathbf{W}^T \mathbf{x}_i - \mathbf{W}^T\mathbf{x}_j ||^2 \mathbf{P}_{ij} + \gamma ||\mathbf{W}||_{21}  \nonumber  \\
\st \quad &  \mathbf{W}^T \mathbf{X} \mathbf{X}^T \mathbf{W} = \mathbf{I} 
\end{align}
Using $\mathbf{L}_{\mathbf{S}} = (\mathbf{I} - \mathbf{S}) (\mathbf{I} - \mathbf{S})^T$, $\mathbf{L}_{\mathbf{P}} = \mathbf{D}_{\mathbf{P}} - (\mathbf{P} + \mathbf{P}^T)/2$ and let $\mathbf{L} = \mathbf{L}_{\mathbf{S}} + \beta\mathbf{L}_{\mathbf{P}}$, the above problem can be rewritten as
\begin{align}\label{opt_w}
\min_{\mathbf{W}} \quad & Tr(\mathbf{W}^T\mathbf{X} \mathbf{L} \mathbf{X}^T \mathbf{W}^T ) + \gamma ||\mathbf{W}||_{21} \\
\st \quad &  \mathbf{W}^T \mathbf{X} \mathbf{X}^T \mathbf{W} = \mathbf{I} \nonumber
\end{align}
Due to the non-smooth regularization, it is hard to obtain the close form solution. We solve it in an iterative way. Given the $t$-th estimation $\mathbf{W}^t$ and denote $\mathbf{D}_{\mathbf{W}^t}$ be a diagonal matrix with the $i$-th diagonal
element as $\frac{1}{2 ||\bm{w}_i^t||^2}$, Eq. \eqref{opt_w} can be rewritten as:
\begin{align}
\min_{\mathbf{W}} \quad & Tr\left(\mathbf{W}^T\mathbf{X} (\mathbf{L} + \gamma \mathbf{D}_{\mathbf{W}^t} ) \mathbf{X}^T \mathbf{W} \right) \\
\st \quad &  \mathbf{W}^T \mathbf{X} \mathbf{X}^T \mathbf{W} = \mathbf{I} \nonumber
\end{align}
The optimal solution of $\mathbf{W}$ are the eigenvectors corresponding to the $c$ smallest eigenvalues of generalized eigen-problem:
\begin{align}\label{g_w}
\mathbf{X} (\mathbf{L} + \gamma \mathbf{D}_{\mathbf{W}^t} ) \mathbf{X}^T \mathbf{W} = \Lambda \mathbf{X} \mathbf{X}^T \mathbf{W} 
\end{align}
where $\Lambda$ is a diagonal matrix whose diagonal elements are eigenvalues. To
get a stable solution of this eigen-problem, the matrices $\mathbf{X} \mathbf{X}^T$ is required to be non-singular which is not true when the number of features is larger than the number of samples. Moreover, the computational complexity of this approach scales as $O(d^3 + nd^2)$, which is costly for high dimensional data. Thus, such solution is less attractive in real world applications. To improve the effectiveness and the efficiency to optimize Eq. \eqref{opt_w}, we further resort to a two steps procedure inspired from \cite{sr}.
\begin{theorem}\label{the_w}
Let $\mathbf{Y} \in \mathcal{R}^{n \times c}$ be a matrix of which each column is an eigenvector of eigen-problem $\mathbf{L} \mathbf{y} = \lambda \mathbf{y}$. If there exists a matrix $\mathbf{W} \in \mathcal{R}^{d \times c}$ such that $\mathbf{X}^T \mathbf{W} = \mathbf{Y}$, then each column of $\mathbf{W}$ is an eigenvector of the generalized eigen-problem $\mathbf{X} \mathbf{L}  \mathbf{X}^T \mathbf{w} = \lambda \mathbf{X} \mathbf{X}^T \mathbf{w} $ with the same eigenvalue $\lambda$.
\begin{proof} With $\mathbf{X}^T \mathbf{W} = \mathbf{Y}$, the following equation holds
\begin{align}
\mathbf{X} \mathbf{L}  \mathbf{X}^T \mathbf{w} = \mathbf{X} \mathbf{L} \mathbf{y} = \mathbf{X} \lambda \mathbf{y} = \lambda \mathbf{X} \mathbf{y} = \lambda \mathbf{X} \mathbf{X}^T \mathbf{w}
\end{align}
Thus, $\mathbf{y}$ is the eigenvector of the generalized eigen-problem $\mathbf{X} \mathbf{L}  \mathbf{X}^T \mathbf{w} = \lambda \mathbf{X} \mathbf{X}^T \mathbf{w} $ with
the same eigenvalue $\lambda$.
\end{proof}
\end{theorem}
Theorem \ref{the_w} shows that instead of solving the generalized eigen-problem in Eq. \eqref{g_w}, $\mathbf{W}$ can be obtained by the following two steps:
\begin{enumerate}
\item Solve the eigen-problem $\mathbf{L} \mathbf{Y} = \Lambda \mathbf{Y}$ to get $\mathbf{Y}$ corresponding to the $c$ smallest eigenvalues;
\item Find $\mathbf{W}$ which satisfies $\mathbf{X}^T \mathbf{W} = \mathbf{Y}$. Since such $\mathbf{W}$ may not exist in real applications, we resort to solve the following optimization problem:
\begin{align}\label{opt_l21}
\min_{\mathbf{W}} \quad ||\mathbf{Y} - \mathbf{X}^T \mathbf{W}||^2 + \gamma ||\mathbf{W}||_{21}
\end{align}
The optimal solution of Eq. \eqref{opt_l21} can also be obtained from routine optimization tools, such as the iterative re-weighted method and the proximal method \cite{liu_slep}.
\end{enumerate}
 
The complete algorithm to solve FSASL is summarized in algorithm \ref{alg_fsasl}.
\begin{algorithm}
	\caption{The optimization algorithm of FSASL}
	\label{alg_fsasl}
	\begin{algorithmic}
	\REQUIRE{The data matrix $\mathbf{X} \in \mathbb{R}^{d \times n}$, the regularization parameters $\alpha$, $\beta$, $\gamma$, $\mu$, the dimension of the transformed data $c$.}
	\REPEAT
	\STATE{For each $i$, update the $i$-th column of $\mathbf{S}$ by solving the problem in Eq. \eqref{opt_s};}
	\STATE{For each $i$, update the $i$-th row of $\mathbf{P}$ using Algorithm \ref{alg_proj};}
	\STATE{Compute the overall graph Laplacian $\mathbf{L}=\mathbf{L}_{\mathbf{S}}+\beta\mathbf{L}_{\mathbf{P}}$;}
	\STATE{Compute $\mathbf{W}$ by Eq. \eqref{g_w} or Eq. \eqref{opt_l21};}
	\UNTIL{Converges}
	\ENSURE{Sort all the $d$ features according to $||\mathbf{w}_i||_2 (i = 1,...,d)$ in descending order and select the top $m$ ranked features.}
	\end{algorithmic}
\end{algorithm}

\subsection{Convergence Analysis}
FSASL is solved in an alternative way, the optimization procedure will monotonically decrease the objective of the problem in Eq. \eqref{opt_fsasl} in each iteration. Since the objective function has lower bounds, such as zero, the above iteration converges. Besides, the experimental results show that it converges fast, the time of iteration is often less than 20.

\subsection{The determination of parameter $\mu$}
Since the parameter $\mu$ is used to control the trade off between the trivial solution ($\mu = 0$) and the uniform distribution ($\mu = \infty$), we would like to keep only top-$k$ neighbors for local manifold structure characterization as the $k$-nn graph \cite{pcan}. We provide an effective method to achieve this. For each sub problem in Eq. \eqref{opt_proj}, the Lagrangian function is 
\begin{align}
\frac{1}{2}|| \mathbf{p}_i^T - \mathbf{a}_i^T ||^2 -  \tau(  \mathbf{p}_i^T \mathbf{1}_n- 1 ) - \eta_i^T \mathbf{p}_i
\end{align}
where $ \tau $ and $ \eta_i $ are the Lagrangian multipliers. According to KKT condition, the optimal value can be obtained by
\begin{align}\label{P}
\mathbf{P}_{ij} = (\mathbf{A}_{ij} + \tau)_{+}
\end{align}
By sorting each row of $\mathbf{A}$ into $\mathbf{B}$ with ascending order,  the following inequality holds
\begin{align}
\left\{ \begin{array}{ll} B_{ik'} + \tau > 0 & \text{for } k' = 1, ..., k \\
						  B_{ik'} + \tau \leq 0 & \text{for }k' = k+1, ..., n
		\end{array} \right.
\end{align}
Considering the simplex constraint on $\mathbf{p}_i^T$, we further get 
\begin{align}\label{tau}
\tau = \frac{1}{k}(1 - \sum_{k'=1}^{k}B_{ik'})
\end{align}
By replacing Eq. \eqref{tau} into Eq. \eqref{P}, the optimal value of $\mathbf{P}$ can be obtained by
\begin{align}\label{optval_P}
\mathbf{P}_{ij} = (\mathbf{A}_{ij} - \frac{1}{k}(1 - \sum_{k'=1}^{k}B_{ik'}))_{+}
\end{align}
 Since $B_{ik'} = -\frac{1}{2\mu} ||\mathbf{Wx}_i - \mathbf{Wx}_{k'} ||^2 = d_{ik'}^{\mathbf{W}}$, for each subproblem we have
\begin{align}
\frac{k}{2}d_{ik}^{\mathbf{W}} - \frac{1}{2}\sum_{k'=1}^{k}d_{ik'}^{\mathbf{W}} < \mu \leq \frac{k}{2}d_{i,k'+1}^{\mathbf{W}} - \frac{1}{2}\sum_{k'=1}^{k}d_{ik'}^{\mathbf{W}}
\end{align}
When $\mu$ satisfies the above inequality for $i$-th example, the corresponding $\mathbf{p}_i^T$ has $k$ non-zero component. Therefore the average non-zero elements in each row of $\mathbf{P}$ is close to $k$ when we set 
\begin{align}\label{compute_mu}
\mu = \frac{1}{n} \sum_{i=1}^{n}\left( \frac{k}{2}d_{i,k'+1}^{\mathbf{W}} - \frac{1}{2}\sum_{k'=1}^{k}d_{ik'}^{\mathbf{W}}\right)
\end{align} 
In this way, the search of parameter $\mu$ can be better handled by searching the neighborhood size $k$, which is more intuitive and easy to tune.

\section{Discussion}

In this section, we discuss some approaches which are closely related to our method.

Zeng and Cheung \cite{llcfs} proposed to integrate feature selection within the regularization of local learning-based clustering (LLCFS), which involves two sub steps:
\begin{enumerate}
\item It constructs the $k$-nearest neighbor graph in the weighted feature space.
\item It performs joint clustering and feature weight learning by solving the following problem
\begin{align}
\min_{\mathbf{Y}, \{\mathbf{W}^{i}, \mathbf{b}^{i} \}_{i=1}^{n}, \mathbf{z}} \quad & \sum_{i=1}^{n}\sum_{c'=1}^{c}\left[ \sum_{\mathbf{x}_j \in \mathcal{N}_{\mathbf{x}_i}} \beta (\mathbf{Y}_{ic'} - \mathbf{x}_j^T \mathbf{W}_{c'}^{i} - \mathbf{b}_{c'}^{i})^2 \right. \nonumber \\
& + \left. (\mathbf{W}_{c'}^{i})^T \diag(\mathbf{z}^{-1}) \mathbf{W}_{c'}^{i}\ \right] \\
\st \quad &  \bm{1}_d^T \bm{z} = 1, \mathbf{z} \geq 0 \nonumber
\end{align}
where $\mathbf{z}$ is the feature weight vector and $\mathcal{N}_{\mathbf{x}_i}$ is the $k$-nearest neighbor of $\mathbf{x}_i$ based on $\mathbf{z}$ weighted features.
\end{enumerate}
Compared with LLCFS, FSASL performs both the global and local structure learning in an adaptive manner, where only local structure is explored by LLCFS. Moreover, LLCFS uses the discrete $k$-nearest graph and does not optimize the same objective function in structure learning and feature search, while FSASL is optimized within a unified framework with the probabilistic neighborhood relationship.

Hou et al. \cite{jelsr2} proposed the joint embedding
learning and sparse regression (JELSR) method, which can be formulated as follows:
\begin{align}\label{opt_jelsr}
\min_{\mathbf{W}, \mathbf{Y}^T \mathbf{Y} = \mathbf{I}}  & \tr(\mathbf{Y}^T \mathbf{L}_2 \mathbf{Y}) + \lambda_1(|| \mathbf{Y} - \mathbf{X}^T \mathbf{W}||^2 + \lambda_2 ||\mathbf{W}||_{21})
\end{align}
Comparing the formulation in Eq. \eqref{opt_fsasl} and Eq. \eqref{opt_jelsr}, the main differences between FSASL and JELSR include: 1) FSASL selects those features to respect both the global and local manifold structure, while JELSR only incorporates the local manifold structure; 2) The local structure in JELSR is based on $k$-nearest neighbor graph, while FSASL learns a probabilistic neighborhood graph, which can be easily refined according the result of feature selection. 3)JELSR iteratively perform spectral embedding for clustering and sparse spectral regression for feature selection as illustrated in Fig. \eqref{ufs_tree}. However, the local structure itself (i.e. $\mathbf{L}_2$) is not changed during iterations. FSASL can adaptively improve both the global and local structure characterization using selected features.

Most recently, Liu et al. \cite{glspfs} proposed a global and local structure preservation framework for feature selection (GLSPFS). It first constructs the pairwise sample similarity matrix $\mathbf{K}$ with Gaussian kernel function to capture the global structure of data, then decompose $\mathbf{K} = \mathbf{Y}\mathbf{Y}^T$. Using $\mathbf{Y}$ as the regression target, GLSPFS solve the following problem:
\begin{align}
\min_{\mathbf{W}} ||\mathbf{Y} - \mathbf{X}^T \mathbf{W}||^2 + \lambda_1 \tr(\mathbf{W}^T \mathbf{X} \mathbf{L}_3 \mathbf{X}^T \mathbf{W}) + \lambda_2 ||\mathbf{W}||_{21}
\end{align}
The main differences between FSASL and GLSPFS include: 1) GLSPFS uses the Gaussian kernel, while FSASL captures the global structure within sparse representation, which is more discriminant; 2) Both the global and local structures (i.e. $\mathbf{K}$ and $\mathbf{L}_3$) in GLSPFS are based on all features, while FSASL refines these structures with selected features.

\begin{table*}[t]
\centering
\caption{Aggregated clustering results measured by Accuracy (\%) of the compared methods.}
\label{res_acc}
\begin{tabular}{l ||c|| c c c c c c c c c c}
    \toprule
    Data Sets & AllFea & LapScore & MCFS & LLCFS & UDFS & NDFS & SPFS & RUFS & JELSR & GLSPFS & FSASL \\
    \hline
    MFEA & 68.73 & \tabincell{c}{ 51.78 \\ $\pm$ 5.51 \\0.00 } & \tabincell{c}{ 51.04 \\ $\pm$ 8.13 \\0.00 } & \tabincell{c}{ 60.38 \\ $\pm$ 8.58 \\0.00 } & \tabincell{c}{ 64.94 \\ $\pm$ 3.32 \\0.00 } & \tabincell{c}{ 67.13 \\ $\pm$ 7.53 \\0.01 } & \tabincell{c}{ \textbf{68.20} \\ \textbf{$\pm$ 9.43 } \\ \textbf{0.22 }} & \tabincell{c}{ 64.58 \\ $\pm$ 7.99 \\0.00 } & \tabincell{c}{ 67.01 \\ $\pm$ 8.37 \\0.01 } & \tabincell{c}{ 61.00 \\ $\pm$ 8.70 \\0.00 } & \tabincell{c}{ \textbf{69.94} \\ \textbf{$\pm$ 7.19 } \\ \textbf{1.00 }} \\ \hline
    USPS49 & 77.70 & \tabincell{c}{ 69.21 \\ $\pm$ 8.95 \\0.00 } & \tabincell{c}{ 84.77 \\ $\pm$ 1.59 \\0.00 } & \tabincell{c}{ 94.96 \\ $\pm$ 1.44 \\0.03 } & \tabincell{c}{ 94.05 \\ $\pm$ 1.13 \\0.00 } & \tabincell{c}{ 68.12 \\ $\pm$ 8.18 \\0.00 } & \tabincell{c}{ 83.43 \\ $\pm$ 6.66 \\0.00 } & \tabincell{c}{ 85.86 \\ $\pm$ 2.58 \\0.00 } & \tabincell{c}{ 95.16 \\ $\pm$ 0.55 \\0.00 } & \tabincell{c}{ 94.75 \\ $\pm$ 0.61 \\0.00 } & \tabincell{c}{ \textbf{95.95} \\ \textbf{$\pm$ 0.48 } \\ \textbf{1.00 }} \\ \hline

  	UMIST & 42.40 & \tabincell{c}{ 36.73 \\ $\pm$ 1.18 \\0.00 } & \tabincell{c}{ 44.46 \\ $\pm$ 3.26 \\0.00 } & \tabincell{c}{ 47.31 \\ $\pm$ 0.83 \\0.00 } & \tabincell{c}{ 48.04 \\ $\pm$ 1.92 \\0.00 } & \tabincell{c}{ 52.80 \\ $\pm$ 2.26 \\0.00 } & \tabincell{c}{ 46.72 \\ $\pm$ 1.70 \\0.00 } & \tabincell{c}{ 50.87 \\ $\pm$ 1.95 \\0.00 } & \tabincell{c}{ 53.52 \\ $\pm$ 1.54 \\0.01 } & \tabincell{c}{ 50.53 \\ $\pm$ 0.59 \\0.00 } & \tabincell{c}{ \textbf{54.92} \\ \textbf{$\pm$ 1.89 } \\ \textbf{1.00 }} \\ \hline
	 
    JAFFE & 71.57 & \tabincell{c}{ 67.62 \\ $\pm$ 8.49 \\0.00 } & \tabincell{c}{ 73.56 \\ $\pm$ 4.83 \\0.00 } & \tabincell{c}{ 64.79 \\ $\pm$ 4.08 \\0.00 } & \tabincell{c}{ 75.48 \\ $\pm$ 1.63 \\0.00 } & \tabincell{c}{ 74.98 \\ $\pm$ 2.15 \\0.00 } & \tabincell{c}{ 73.93 \\ $\pm$ 2.85 \\0.00 } & \tabincell{c}{ 75.75 \\ $\pm$ 2.53 \\0.00 } & \tabincell{c}{ 77.77 \\ $\pm$ 1.87 \\0.00 } & \tabincell{c}{ 75.46 \\ $\pm$ 1.61 \\0.00 } & \tabincell{c}{ \textbf{79.29} \\ \textbf{$\pm$ 2.24 } \\ \textbf{1.00 }} \\ \hline
	 
    AR & 30.26 & \tabincell{c}{ 25.29 \\ $\pm$ 2.89 \\0.00 } & \tabincell{c}{ 29.05 \\ $\pm$ 1.19 \\0.00 } & \tabincell{c}{ \textbf{34.22} \\ \textbf{$\pm$ 2.70 } \\ \textbf{0.05 }} & \tabincell{c}{ 30.87 \\ $\pm$ 0.35 \\0.00 } & \tabincell{c}{ 32.34 \\ $\pm$ 1.52 \\0.00 } & \tabincell{c}{ 31.06 \\ $\pm$ 2.14 \\0.00 } & \tabincell{c}{ 34.84 \\ $\pm$ 1.90 \\0.04 } & \tabincell{c}{ 34.19 \\ $\pm$ 2.52 \\0.02 } & \tabincell{c}{ 34.12 \\ $\pm$ 1.60 \\ 0.00} & \tabincell{c}{ \textbf{36.11} \\ \textbf{$\pm$ 0.75 } \\ \textbf{1.00 }} \\ \hline
	
    COIL & 59.17 & \tabincell{c}{ 45.60 \\ $\pm$ 6.16 \\0.00 } & \tabincell{c}{ 51.50 \\ $\pm$ 5.38 \\0.00 } & \tabincell{c}{ 50.84 \\ $\pm$ 3.76 \\0.00 } & \tabincell{c}{ 48.40 \\ $\pm$ 16.89 \\0.00 } & \tabincell{c}{ 52.22 \\ $\pm$ 6.33 \\0.00 } & \tabincell{c}{ 56.94 \\ $\pm$ 3.43 \\0.00 } & \tabincell{c}{ 59.20 \\ $\pm$ 3.28 \\0.00 } & \tabincell{c}{ 59.53 \\ $\pm$ 4.01 \\0.03 } & \tabincell{c}{ 57.96 \\ $\pm$ 2.27 \\0.00 } & \tabincell{c}{ \textbf{60.93} \\ \textbf{$\pm$ 2.50 } \\ \textbf{1.00 }} \\ \hline
	
    LUNG & 72.46 & \tabincell{c}{ 58.97 \\ $\pm$ 5.24 \\0.00 } & \tabincell{c}{ 70.42 \\ $\pm$ 3.41 \\0.00 } & \tabincell{c}{ 71.58 \\ $\pm$ 5.85 \\0.00 } & \tabincell{c}{ 65.46 \\ $\pm$ 3.88 \\0.00 } & \tabincell{c}{ 75.52 \\ $\pm$ 1.57 \\0.00 } & \tabincell{c}{ 73.49 \\ $\pm$ 3.43 \\0.00 } & \tabincell{c}{ 77.35 \\ $\pm$ 2.62 \\0.00 } & \tabincell{c}{ 77.86 \\ $\pm$ 3.12 \\0.00 } & \tabincell{c}{ 77.83 \\ $\pm$ 2.70 \\0.00 } & \tabincell{c}{ \textbf{81.93} \\ \textbf{$\pm$ 1.63 } \\ \textbf{1.00 }} \\ \hline
	TOX & 43.65 & \tabincell{c}{ 40.25 \\ $\pm$ 0.65 \\0.00 } & \tabincell{c}{ 43.10 \\ $\pm$ 1.86 \\0.00 } & \tabincell{c}{ 39.28 \\ $\pm$ 0.49 \\0.00 } & \tabincell{c}{ 47.14 \\ $\pm$ 0.75 \\0.00 } & \tabincell{c}{ 38.28 \\ $\pm$ 1.64 \\0.00 } & \tabincell{c}{ 39.93 \\ $\pm$ 1.13 \\0.00 } & \tabincell{c}{ 49.17 \\ $\pm$ 0.83  \\ 0.00 } & \tabincell{c}{ 43.96 \\ $\pm$ 1.56 \\0.00 } & \tabincell{c}{ 47.38 \\ $\pm$ 1.93 \\0.00 } & \tabincell{c}{ \textbf{50.12} \\ \textbf{$\pm$ 0.67} \\ \textbf{1.00} } \\ \hline
	Average & 58.24 & 49.43 & 55.98 & 57.92 & 59.29 & 56.67 & 59.21 & 62.2 & 63.62 & 62.38 & \textbf{66.15} \\
    \bottomrule
\end{tabular}
\end{table*}

\begin{table*}[t]
\centering
\caption{Aggregated clustering results measured by Normalized Mutual Information (\%) of the compared methods.}
\label{res_nmi}
\begin{tabular}{l c c c c c c c c c c c}
    \toprule
    Data Sets & AllFea & LapScore & MCFS & LLCFS & UDFS & NDFS & SPFS & RUFS & JELSR & GLSPFS & FSASL \\
    \midrule
    MFEA & 70.33 & \tabincell{c}{ 53.74 \\ $\pm$ 4.77 \\0.00 } & \tabincell{c}{ 54.72 \\ $\pm$ 9.14 \\0.00 } & \tabincell{c}{ 52.77 \\ $\pm$ 9.76 \\0.00 } & \tabincell{c}{ 54.19 \\ $\pm$ 3.83 \\0.00 } & \tabincell{c}{ 64.97 \\ $\pm$ 7.54 \\0.03 } & \tabincell{c}{ \textbf{64.92} \\ \textbf{$\pm$ 8.27 } \\ \textbf{0.11 }} & \tabincell{c}{ 63.98 \\ $\pm$ 7.22 \\0.00 } & \tabincell{c}{ \textbf{64.51} \\ \textbf{$\pm$ 9.07 } \\ \textbf{0.06 }} & \tabincell{c}{ 59.26 \\ $\pm$ 7.59 \\0.00 } & \tabincell{c}{ \textbf{66.70} \\ \textbf{$\pm$ 6.71 } \\ \textbf{1.00 }} \\ \hline
    USPS49 & 23.51 & \tabincell{c}{ 15.88 \\ $\pm$ 17.98 \\0.00 } & \tabincell{c}{ 63.14 \\ $\pm$ 1.05 \\0.00 } & \tabincell{c}{ 72.03 \\ $\pm$ 5.56 \\0.03 } & \tabincell{c}{ 68.12 \\ $\pm$ 4.46 \\0.00 } & \tabincell{c}{ 62.27 \\ $\pm$ 9.62 \\0.00 } & \tabincell{c}{ 68.10 \\ $\pm$ 16.66 \\0.00 } & \tabincell{c}{ 71.73 \\ $\pm$ 7.23 \\0.00 } & \tabincell{c}{ 72.28 \\ $\pm$ 2.24 \\0.00 } & \tabincell{c}{ 70.43 \\ $\pm$ 2.57 \\0.00 } & \tabincell{c}{ \textbf{75.88} \\ \textbf{$\pm$ 2.28 } \\ \textbf{1.00 }} \\ \hline
	 
    UMIST & 64.15 & \tabincell{c}{ 55.57 \\ $\pm$ 2.32 \\0.00 } & \tabincell{c}{ 63.46 \\ $\pm$ 4.93 \\0.00 } & \tabincell{c}{ 63.42 \\ $\pm$ 1.42 \\0.00 } & \tabincell{c}{ 65.19 \\ $\pm$ 2.96 \\0.00 } & \tabincell{c}{ 71.19 \\ $\pm$ 2.77 \\0.01 } & \tabincell{c}{ 64.90 \\ $\pm$ 3.06 \\0.00 } & \tabincell{c}{ 68.19 \\ $\pm$ 2.61 \\0.00 } & \tabincell{c}{ 71.33 \\ $\pm$ 2.06 \\0.00 } & \tabincell{c}{ 69.16 \\ $\pm$ 0.97 \\0.00 } & \tabincell{c}{ \textbf{72.39} \\ \textbf{$\pm$ 2.39 } \\ \textbf{1.00 }} \\ \hline
	JAFFE & 81.52 & \tabincell{c}{ 77.28 \\ $\pm$ 8.98 \\0.00 } & \tabincell{c}{ 79.04 \\ $\pm$ 5.88 \\0.00 } & \tabincell{c}{ 66.97 \\ $\pm$ 3.47 \\0.00 } & \tabincell{c}{ 84.25 \\ $\pm$ 1.74 \\0.00 } & \tabincell{c}{ 82.53 \\ $\pm$ 3.49 \\0.00 } & \tabincell{c}{ 80.01 \\ $\pm$ 3.06 \\0.00 } & \tabincell{c}{ 82.00 \\ $\pm$ 3.56 \\0.00 } & \tabincell{c}{ 85.23 \\ $\pm$ 3.31 \\0.00 } & \tabincell{c}{ 83.20 \\ $\pm$ 3.17 \\0.00 } & \tabincell{c}{ \textbf{86.42} \\ \textbf{$\pm$ 3.34 } \\ \textbf{1.00 }} \\ \hline
	AR & 65.48 & \tabincell{c}{ 63.59 \\ $\pm$ 2.36 \\0.00 } & \tabincell{c}{ 66.41 \\ $\pm$ 0.85 \\0.00 } & \tabincell{c}{ 69.01 \\ $\pm$ 1.45 \\0.01 } & \tabincell{c}{ 67.49 \\ $\pm$ 0.27 \\0.00 } & \tabincell{c}{ 67.89 \\ $\pm$ 0.89 \\0.00 } & \tabincell{c}{ 66.94 \\ $\pm$ 1.11 \\0.00 } & \tabincell{c}{ 69.54 \\ $\pm$ 1.10 \\0.01 } & \tabincell{c}{ 69.02 \\ $\pm$ 1.32 \\0.00 } & \tabincell{c}{ 69.44 \\ $\pm$ 0.84 \\ 0.00} & \tabincell{c}{ \textbf{70.78} \\ \textbf{$\pm$ 0.63 } \\ \textbf{1.00 }} \\ \hline
	COIL & 75.58 & \tabincell{c}{ 62.21 \\ $\pm$ 4.98 \\0.00 } & \tabincell{c}{ 66.19 \\ $\pm$ 6.78 \\0.00 } & \tabincell{c}{ 64.04 \\ $\pm$ 4.34 \\0.00 } & \tabincell{c}{ 44.27 \\ $\pm$ 12.61 \\0.00 } & \tabincell{c}{ 56.29 \\ $\pm$ 6.91 \\0.00 } & \tabincell{c}{ 69.91 \\ $\pm$ 4.38 \\0.00 } & \tabincell{c}{ 70.54 \\ $\pm$ 4.48 \\0.00 } & \tabincell{c}{ 71.37 \\ $\pm$ 4.97 \\0.00 } & \tabincell{c}{ 69.89 \\ $\pm$ 4.00 \\0.00 } & \tabincell{c}{ \textbf{72.93} \\ \textbf{$\pm$ 4.44 } \\ \textbf{1.00 }} \\ \hline
	LUNG & 60.37  & \tabincell{c}{ 50.14 \\ $\pm$ 4.13 \\0.00 } & \tabincell{c}{ 55.68 \\ $\pm$ 2.31 \\0.00 } & \tabincell{c}{ 60.12 \\ $\pm$ 4.65 \\0.00 } & \tabincell{c}{ 54.88 \\ $\pm$ 4.21 \\0.00 } & \tabincell{c}{ 60.57 \\ $\pm$ 1.54 \\0.00 } & \tabincell{c}{ 61.75 \\ $\pm$ 3.32 \\0.00 } & \tabincell{c}{ 65.47 \\ $\pm$ 1.87 \\0.00 } & \tabincell{c}{ 63.54 \\ $\pm$ 2.94 \\0.00 } & \tabincell{c}{ 63.50 \\ $\pm$ 2.99 \\0.00 } & \tabincell{c}{ \textbf{66.78} \\ \textbf{$\pm$ 1.72 } \\ \textbf{1.00 }} \\ \hline  
  TOX & 15.87 & \tabincell{c}{ 10.92 \\ $\pm$ 0.68 \\0.00 } & \tabincell{c}{ 16.53 \\ $\pm$ 2.68 \\0.00 } & \tabincell{c}{ 9.68 \\ $\pm$ 0.75 \\0.00 } & \tabincell{c}{ 22.16 \\ $\pm$ 1.36 \\0.00 } & \tabincell{c}{ 9.07 \\ $\pm$ 1.87 \\0.00 } & \tabincell{c}{ 10.13 \\ $\pm$ 1.03 \\0.00 } & \tabincell{c}{ 25.79 \\ $\pm$ 1.60 \\0.00 } & \tabincell{c}{ 17.46 \\ $\pm$ 3.36 \\0.00 } & \tabincell{c}{ 23.49 \\ $\pm$ 2.77 \\0.00 } & \tabincell{c}{ \textbf{27.37} \\ \textbf{$\pm$ 1.62 } \\ \textbf{1.00 }} \\ \hline
	Average & 57.10 & 48.67 & 58.14 & 57.26 & 57.56 & 59.35 & 60.83 & 64.65 & 64.34 & 63.55 & 67.41 \\
  \bottomrule
\end{tabular}
\end{table*}

\section{Experiments}
In this section, we conduct extensive experiments to evaluate the performance of the proposed FSASL for the task of unsupervised feature selection.

\subsection{Data Sets}
The experiments are conducted on 8 publicly available datasets, including handwritten and spoken digit/letter recognition data sets (i.e., MFEA from UCI reporsitory and USPS49 \cite{llcfs} which is a two class subset of USPS), three face image data sets (i.e., UMIST \cite{jelsr2}, JAFFE \cite{ndfs}, AR \cite{ldmgi}), one object data set (i.e. COIL \cite{mcfs}) and two biomedical data sets (i.e., LUNG \cite{rfs_l21}, TOX \cite{asu_data}). The
details of these benchmark data sets are summarized in Table \ref{dataset}.

\begin{table}[h]
\centering
\caption{Summary of the benchmark data sets and the number of selected features}
\begin{tabular}{|c||c|c|c|c|}
\hline
Data Sets & sample & feature & class & selected features\\
\hline
\hline
MFEA & 2000 & 240 & 10 & $[5, 10, \ldots, 50]$\\ \hline 
USPS49 & 1673 & 256 & 2 & $[5, 10, \ldots, 50]$\\ \hline
UMIST & 575 & 644 & 20 & $[5, 10, \ldots, 50]$\\ \hline
JAFFE & 213 & 676 & 10 & $[5, 10, \ldots, 50]$\\ \hline
AR & 840 & 768 & 120 & $[5, 10, \ldots, 50]$\\ \hline
COIL & 1440 & 1024 & 20 & $[5, 10, \ldots, 50]$\\ \hline
LUNG & 203 & 3312 & 5 & $[10, 20, \ldots, 150]$\\ \hline
TOX & 171 & 5748 & 4 & $[10, 20, \ldots, 150]$\\ \hline
\end{tabular}
\label{dataset}
\end{table}

\subsection{Experiment Setup}
To validate the effectiveness of our proposed FSASL\footnote{For the purpose of reproducibility, we provide the code at \url{https://github.com/csliangdu/FSASL}} , we compare it with one baseline (i.e., AllFea) and states-of-the-art unsupervised feature selection methods, 
\begin{itemize}
\item LapScore\footnote{\url{http://www.cad.zju.edu.cn/home/dengcai/Data/code/LaplacianScore.m}} \cite{lapscore}, which evaluates the features according to their ability of locality preserving of the data manifold structure.
\item MCFS\footnote{\url{http://www.cad.zju.edu.cn/home/dengcai/Data/code/MCFS_p.m}} \cite{mcfs}, which selects the features by adopting spectral regression with $\ell_1$-norm regularization.
\item LLCFS \cite{llcfs}, which incorporates the relevance of each feature into the built-in regularization of the local learning-based clustering algorithm.
\item UDFS\footnote{\url{http://www.cs.cmu.edu/~yiyang/UDFS.rar}} \cite{udfs}, which exploits local discriminative information and feature correlations simultaneously.
\item NDFS\footnote{\url{https://sites.google.com/site/zcliustc}} \cite{ndfs}, which selects features by a joint framework of nonnegative spectral analysis and $\ell_{2,1}$-norm regularized regression.
\item SPFS\footnote{\url{https://sites.google.com/site/alanzhao}} \cite{zhao2013similarity}, which selects a feature subset with which the pairwise similarity between high dimensional samples can be maximally preserved.
\item RUFS\footnote{\url{https://sites.google.com/site/qianmingjie}} \cite{qian2013robust}, which performs robust clustering and robust feature selection simultaneously to select the most important and discriminative features.
\item JELSR\footnote{\url{http://www.escience.cn/people/chenpinghou}} \cite{jelsr,jelsr2}, which joins embedding learning with sparse regression to perform feature selection.
\item GLSPFS\footnote{We also use the implementation provided by the authors.} \cite{glspfs}, which integrates both global pairwise sample similarity and local geometric data structure to conduct feature
selection.
\end{itemize}

There are some parameters to be set in advance. For all the feature selection algorithms except SPFS, we set $k = 5$ for all the datasets to specify the size of neighborhoods \cite{mcfs,cgssl}. The weight of $k$-nn graph for LapScore and MCFS, and the pairwise similarity for SPFS and GLSPFS is based on the Gaussian kernel, where the kernel width is searched from the grid $\{2^{-3} , 2^{-2} , \ldots, 2^{3} \}\delta_0 $, where $\delta_0$ is the mean distance between any two data examples. For GLSPFS, we report the best results among three local manifold models, that is locality preserving projection (LPP), LLE and local tangent space alignment (LTSA) as in \cite{glspfs}. For LLCFS, UDFS, NDFS, RUFS, JELSR, GLSPFS and FSASL, the regularization parameters are searched from the grid $\{10^{-5}, 10^{-4}, \ldots, 10^{5}\}$. And the regularization parameter for $\gamma$ is searched from the grid $\{0.001, 0.005, 0.01, 0.05, 0.1\}\gamma_{max}$, where $\gamma_{max}$ is automatically computed from SLEP \cite{liu_slep}. For FSASL, $\mu$ is determined by Eq. \eqref{compute_mu} with $k=5$ and $c$ is set to be the true number of classes. To fairly compare different unsupervised feature selection algorithms, we tune the parameters for all methods by the grid-search strategy \cite{qian2013robust,glspfs}.

With the selected features, we evaluate the performance in terms of $k$-means clustering by two widely used metrics, i.e., Accuracy (ACC) and Normalized Mutual Information (NMI). The results of $k$-means clustering depend on the initialization. For all the compared algorithms with different parameters and different number of selected features, we first repeat the clustering 20 times with random initialization and record the average results.

\begin{figure*}[t]
	\centering
	\subfloat[]{\label{fig:jaffe_para1} \includegraphics[width=0.32\textwidth]{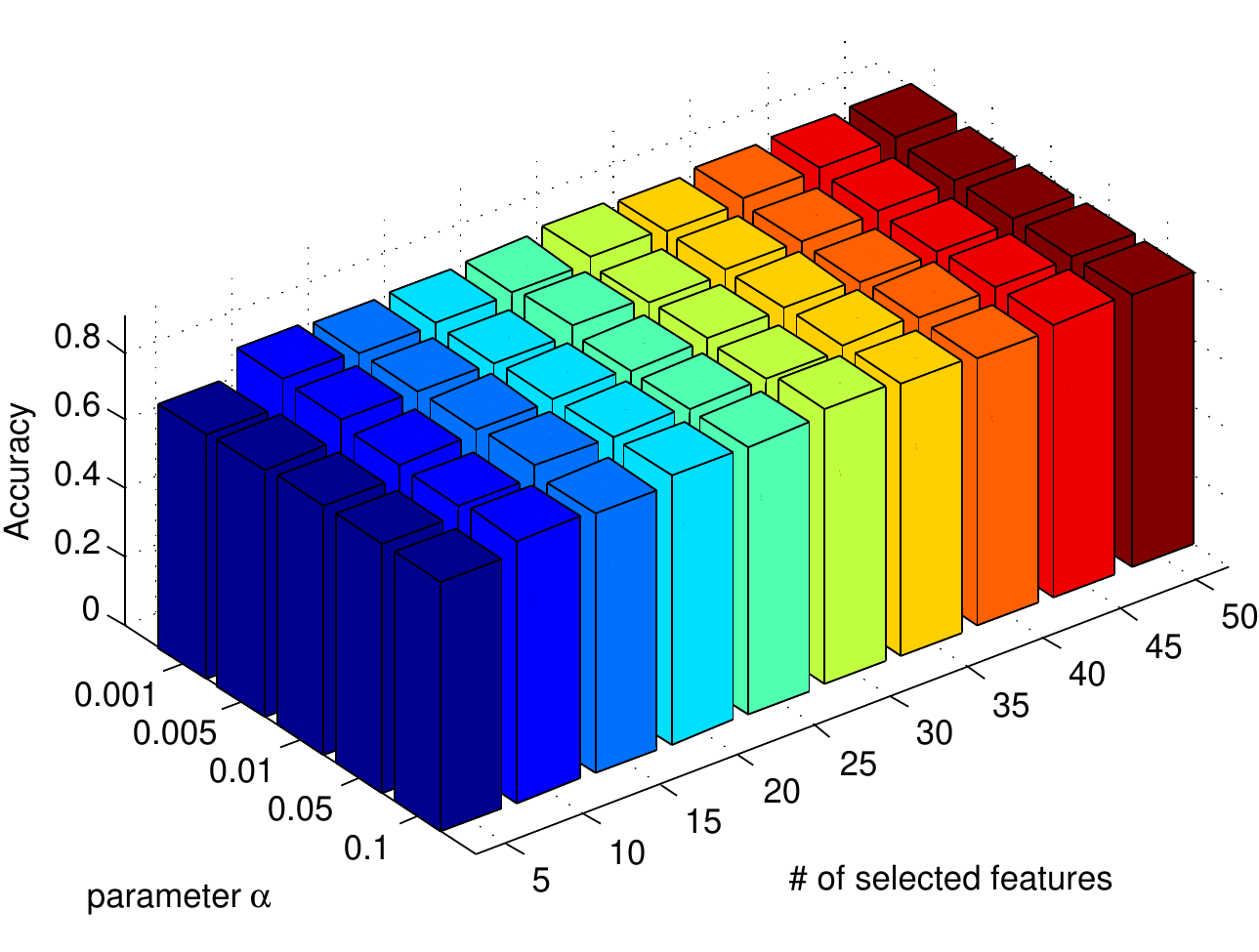}}  
	\subfloat[]{\label{fig:jaffe_para2} \includegraphics[width=0.32\textwidth]{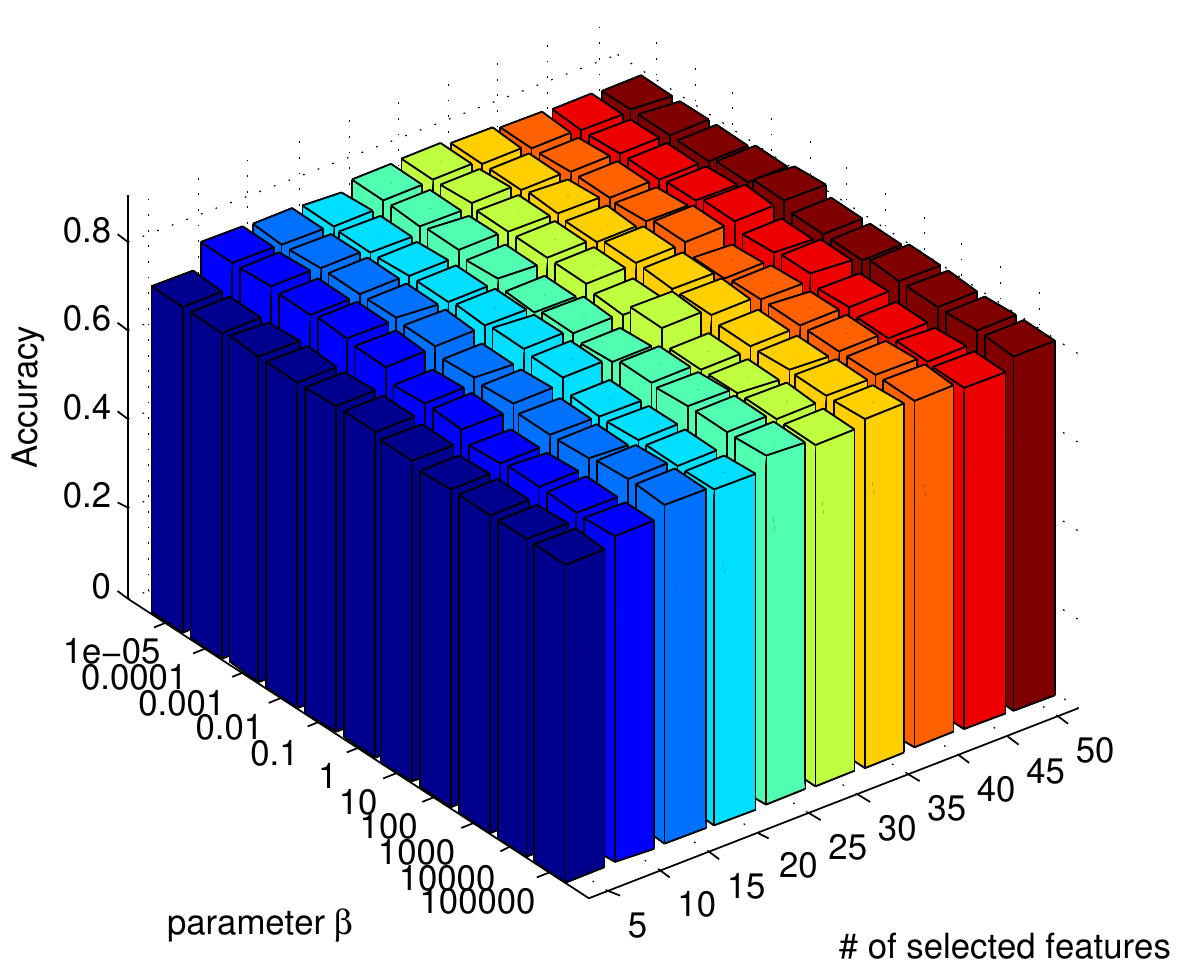}} 
	\subfloat[]{\label{fig:jaffe_para3} \includegraphics[width=0.32\textwidth]{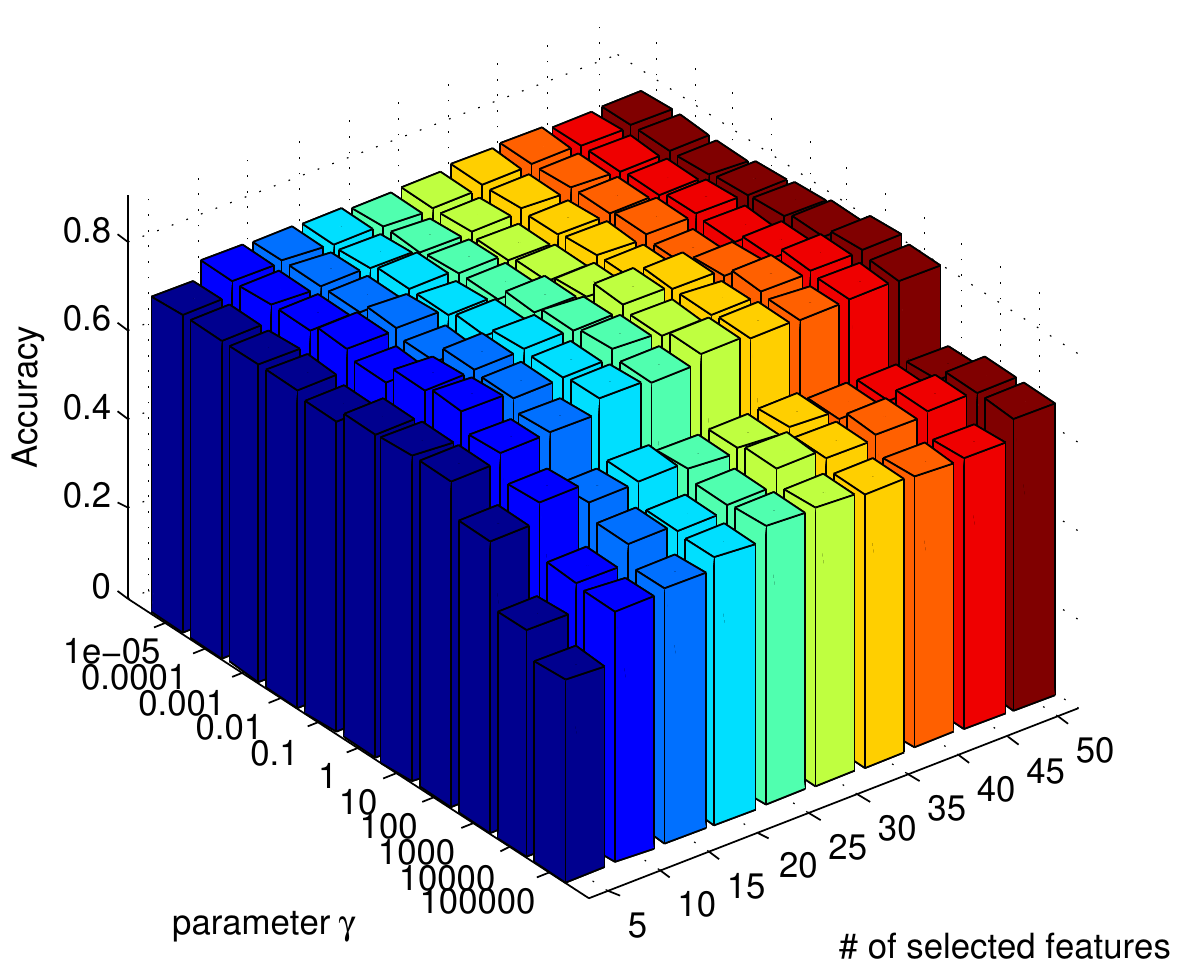}} 
	\\
	\subfloat[]{\label{fig:tox_para1} \includegraphics[width=0.32\textwidth]{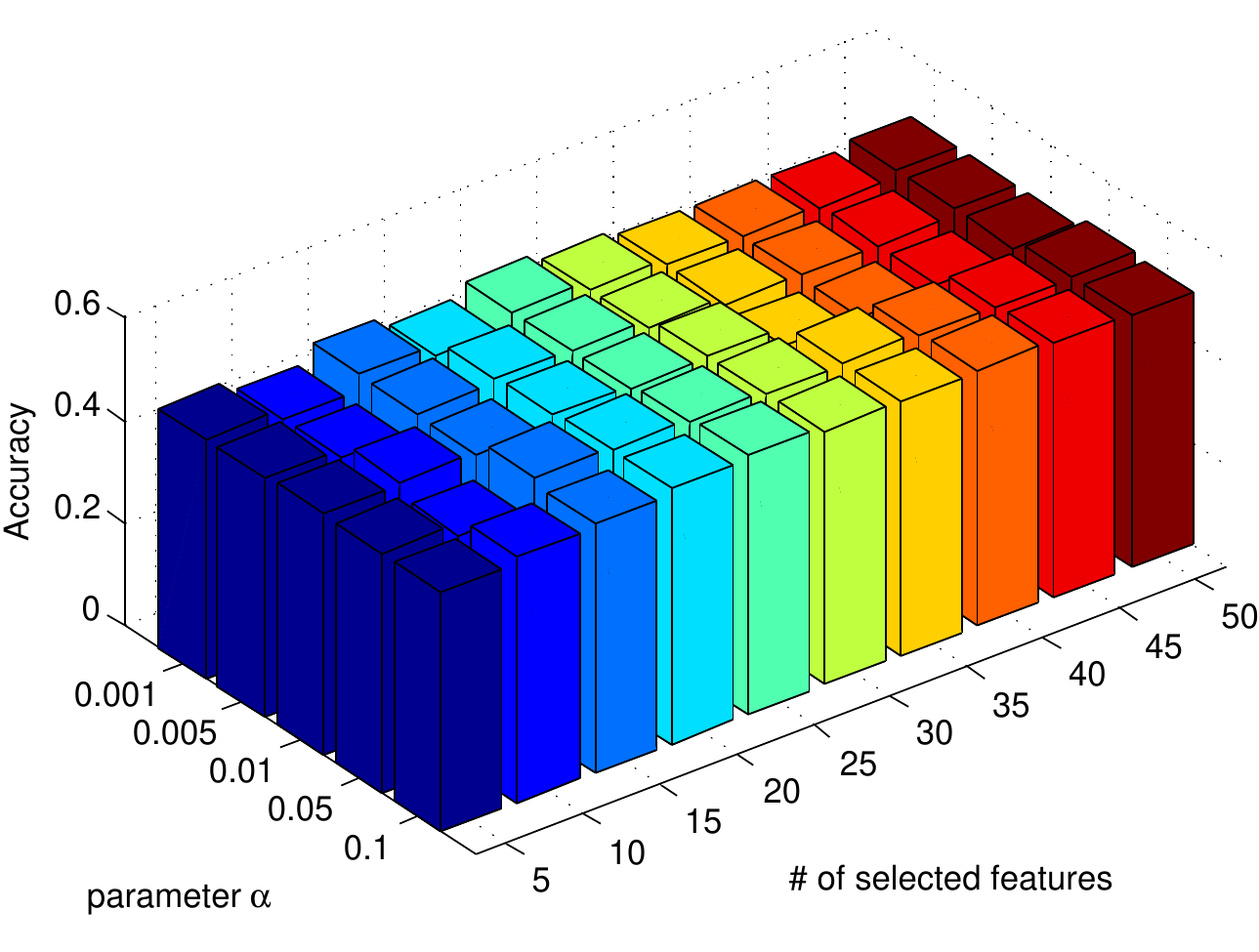}}  
	\subfloat[]{\label{fig:tox_para2} \includegraphics[width=0.32\textwidth]{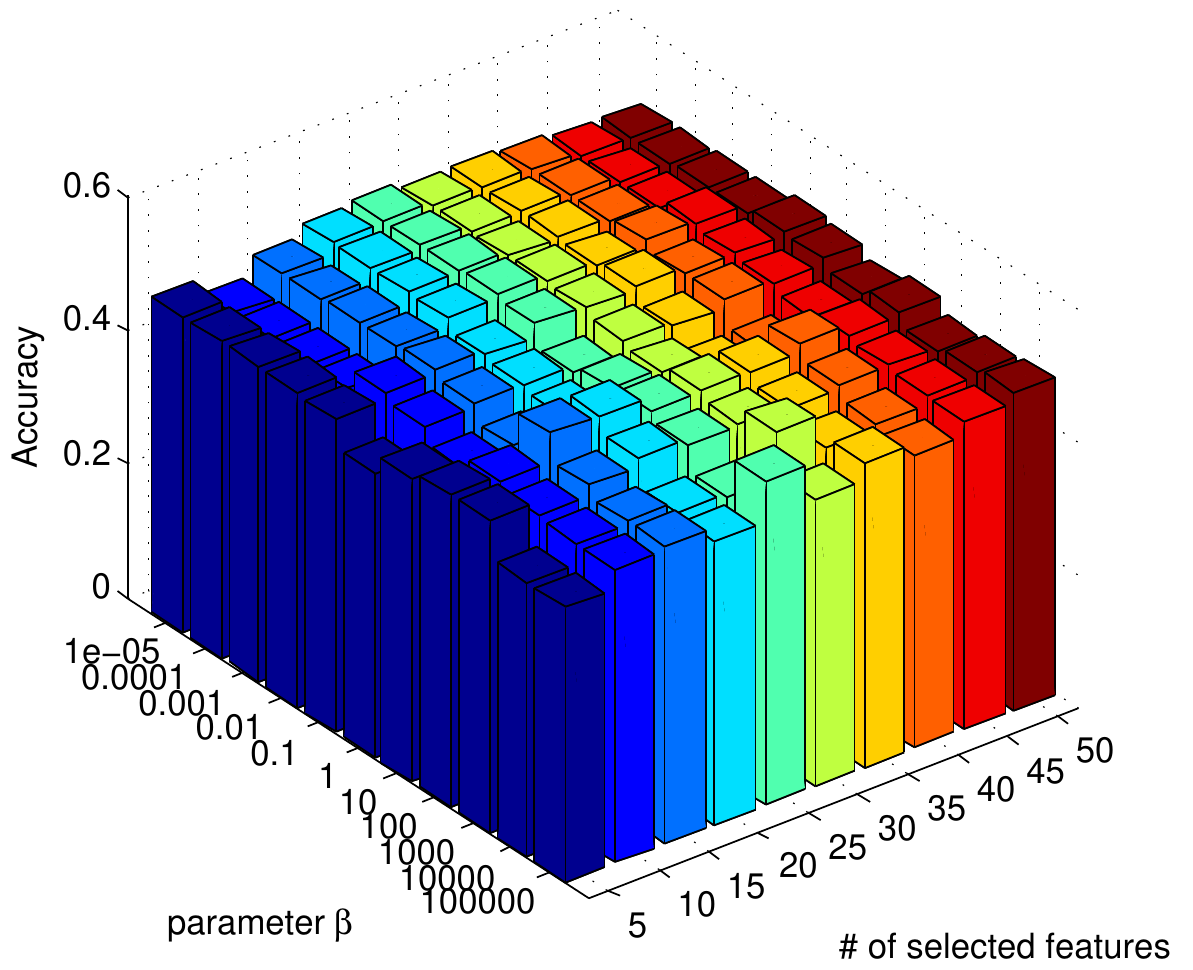}} 
	\subfloat[]{\label{fig:tox_para3} \includegraphics[width=0.32\textwidth]{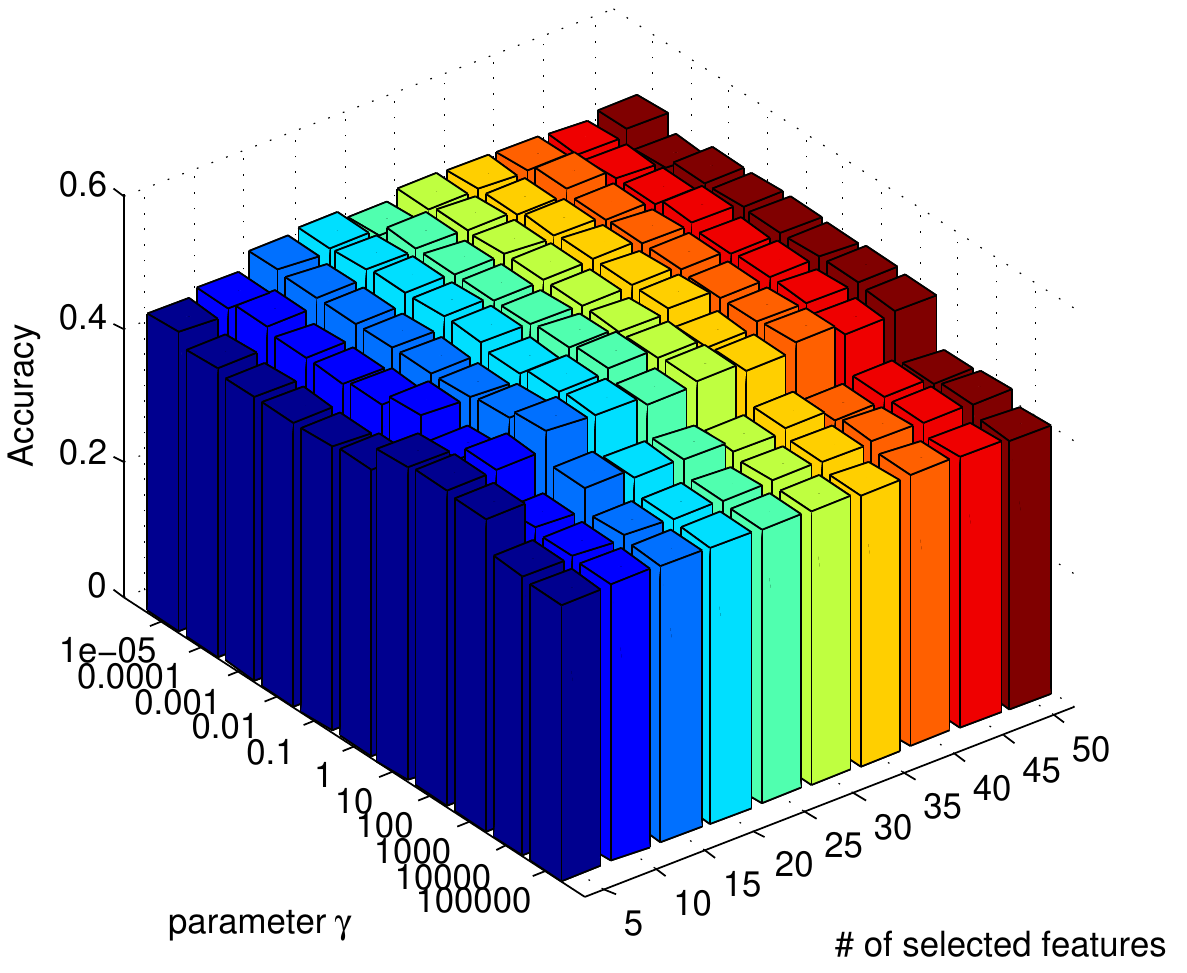}} 
	\caption{Clustering accuracy w.r.t. different parameters on JAFFE (a-c) and TOX (d-f).}
	\label{fig:fsasl_para}
\end{figure*}
\subsection{Clustering with Selected Features}
 Since the optimal number of selected features is unknown in advance, to better evaluate the performance of unsupervised feature selection algorithms, we finally report the averaged results over different number of selected features (the range of selected features for each data set can be found in Table \ref{dataset}) with standard derivation. For all the algorithms (except for AllFea), we also report its $p$-value by the paired $t$-test against the best results. The best one and those having no significant difference $(p > 0.05)$ from the best one are marked in bold.

The clustering results in terms of ACC and NMI are reported in Table \ref{res_acc} and Table \ref{res_nmi}, respectively. For different feature selection algorithms, the results in each cell of Table \ref{res_acc} and \ref{res_nmi} are the mean $\pm$ standard deviation and the $p$-value. The last row of Table \ref{res_acc} and Table \ref{res_nmi} shows the averaged results of all the algorithms over the 8 datasets. 

Compared with clustering using all features, these unsupervised feature selection algorithms not only can largely reduce the number of features facilitating the latter learning process, but can also often improve the clustering performance. In particular, our method FSASL achieves $13.6\%$ and $18.1\%$ improvement in terms of accuracy and NMI respectively with less than $10\%$ features. These results can well demonstrate the effectiveness and efficiency of unsupervised feature selection algorithm. It can also be observed that FSASL consistently produces better performance than the other nine feature selection algorithms, and the improvement is in the range from $4.77\%$ to $38.5\%$ in terms of clustering accuracy and from $3.98\%$ to $33.8\%$ in terms of NMI. This can be mainly explained by the following reasons. First, both global and local structure are used to guide the search of relevant features. Second, the structure learning and feature selection are integrated into a unified framework. Third, both the global and local structures can be adaptively updated using the results of selected features.

\subsection{Effect of Adaptive Structure Learning}
Here, we investigate the effect of adaptive structure learning by empirically answering the following questions:
\begin{enumerate}
\item What kind of structure should be captured and preserved by the selected features, either global or local or both of these structures?
\item Does the adaptive structure learning lead to select more informative features?
\end{enumerate}

\begin{figure}[h]
	\centering
	\includegraphics[width=0.48\textwidth]{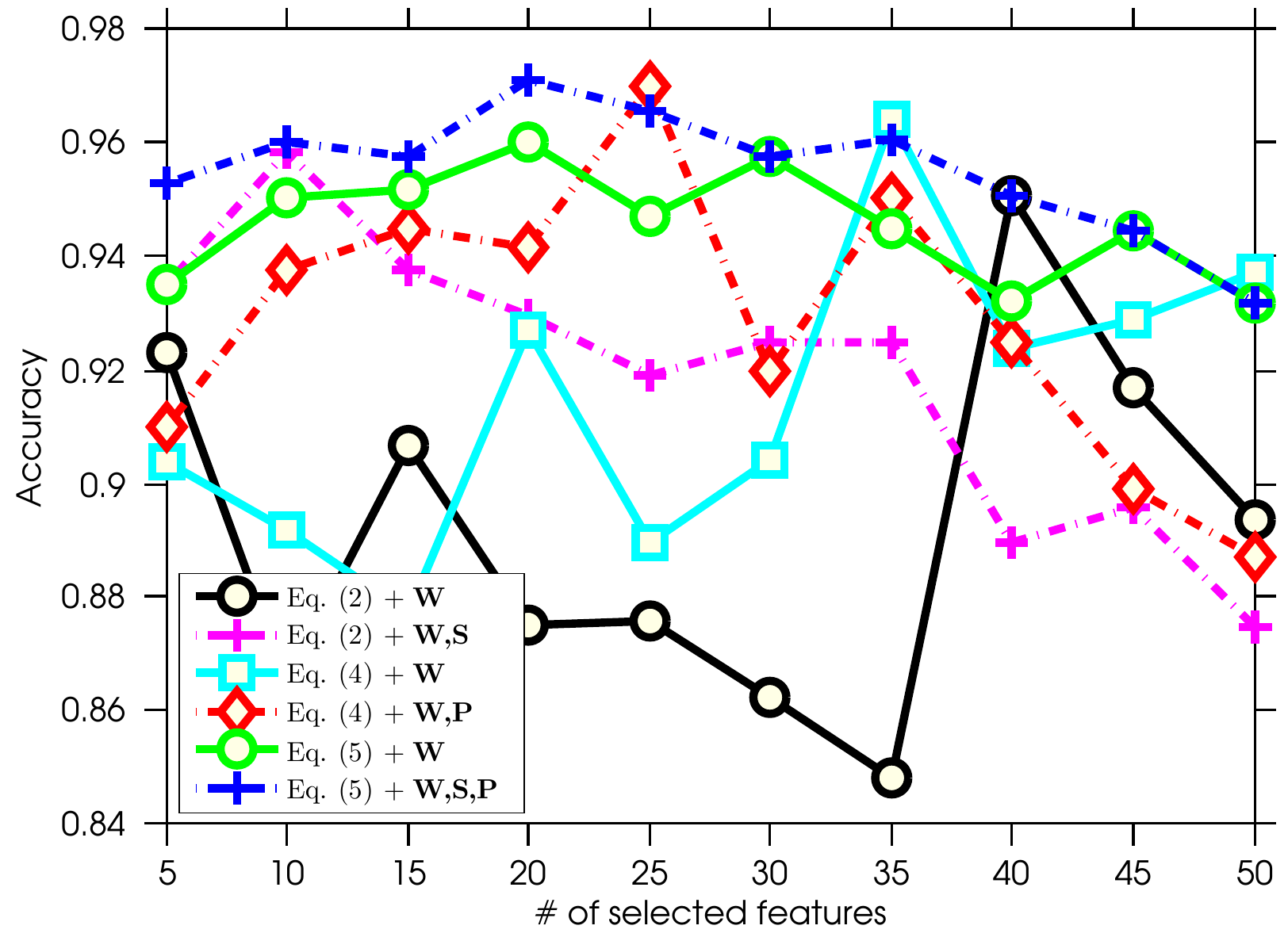}
	\caption{Clustering accuracy w.r.t. 6 different settings of FSASL on USPS200.}
	\label{fig:fsasl_usps_acc}
\end{figure}

\begin{figure}[h]
	\centering
	\includegraphics[width=0.48\textwidth]{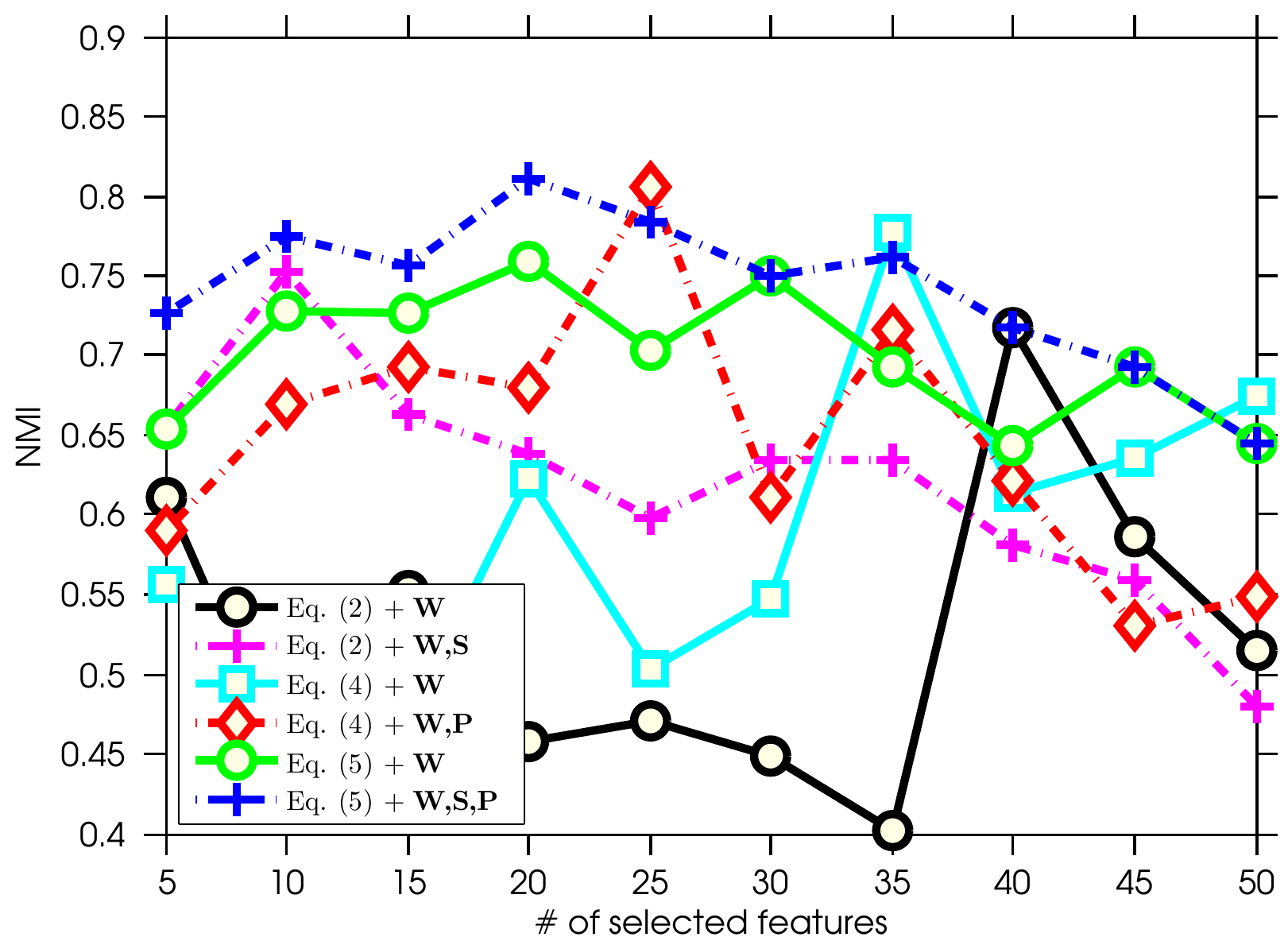}
	\caption{Clustering NMI w.r.t. 6 different settings of FSASL on USPS200.}
	\label{fig:fsasl_usps_nmi}
\end{figure}

\begin{table}[h]
\centering
\caption{Aggregated clustering results ($\%$) of 6 different settings of FSASL on USPS200.}
\begin{tabular}{|c|c||c|c|}
\hline
Problem & Variables & ACC & NMI\\
\hline
\hline
Eq. \eqref{global_asl} & $\mathbf{W}$ & 89.17 $\pm$ 3.22 &  52.01 $\pm$ 9.69 \\ \hline
Eq. \eqref{global_asl} & $\mathbf{W}, \mathbf{S}$ & 91.90 $\pm$ 2.51 & 61.95 $\pm$ 7.21  \\ \hline
\hline
Eq. \eqref{local_asl} & $\mathbf{W}$ & 91.48 $\pm$ 2.62 &  59.10 $\pm$ 9.31\\ \hline
Eq. \eqref{local_asl} & $\mathbf{W}, \mathbf{P}$ & 92.86  $\pm$ 2.53 & 64.65 $\pm$ 8.30 \\ \hline
\hline
Eq. \eqref{opt_fsasl} & $\mathbf{W}$ & 94.65 $\pm$ 1.24 & 69.94 $\pm$ 4.22 \\ \hline
Eq. \eqref{opt_fsasl} & $\mathbf{W}, \mathbf{S}, \mathbf{P}$ & 95.53 $\pm$ 1.10 & 74.20 $\pm$ 4.83 \\ \hline
\end{tabular}
\label{table:fsasl_usps_200}
\end{table}

We conduct different settings of FSASL on USPS200, which consists the first 100 samples in USPS49. We solve the optimization problem in Eq. \eqref{global_asl}, Eq. \eqref{local_asl} and Eq. \eqref{opt_fsasl}, which uses global, local, and both global and local structures, respectively. We also distinguish these problems with and without adaptive structure learning. Thus, we have 6 settings in total. Figure \ref{fig:fsasl_usps_acc} and Figure \ref{fig:fsasl_usps_nmi} show the results of these different settings with different number of selected features. The aggregated result over different number of selected features is also provided in Table \ref{table:fsasl_usps_200}.

From these results, we can see that: 1) The exploitation of both global and local  structures (i.e., Eq. \eqref{opt_fsasl} + $\mathbf{W}$) outperform another two alternatives with only global (i.e., Eq. \eqref{global_asl} + $\mathbf{W}$) or local (i.e., Eq. \eqref{local_asl} + $\mathbf{W}$) structure. It validates that the integration of both global and local structure is better than the single one. 2) With the update of structure learning (i.e., Eq. \eqref{global_asl} + $\mathbf{W,S}$, Eq. \eqref{local_asl} + $\mathbf{W,P}$ and Eq. \eqref{opt_fsasl} + $\mathbf{W,S,P}$ ) is better than their counterparts without adaptive structure learning respectively. It shows that the adaptive learning in either global and/or local structure learning can further improve the result of feature selection.

\subsection{Parameter Sensitivity}
We investigate the sensitivity with respect to the regularization parameters $\alpha$, $\beta$ and $\gamma$. When we vary the value of one parameter, we keep the other parameters fixed at the optimal value. We plot the clustering accuracy with respect to these parameters on JAFFE and TOX in Figure \ref{fig:fsasl_para}. The experimental results show that our method is not very sensitive to $\alpha$, $\beta$ and $\gamma$ with wide ranges. However, the performance is relatively sensitive to the number of selected features, which is still an open problem.

%

\section{Conclusion}
In this paper, we proposed a novel unsupervised feature selection method to simultaneously perform feature selection and the structure learning. In our new method, the global structure learning and feature selection are integrated within the framework of sparse representation; the local structure learning and feature selection are incorporated into the probabilistic neighborhood relationship learning framework. By combining both the global and local structure learning and feature selection, our method can boost both these two essential tasks, i.e., structure learning and feature selection, by using the result of the other task. We derive an efficient algorithm to optimize the proposed method and discuss the connections between our method and other feature selection methods. Extensive experiments have been conducted on real-world benchmark data sets to demonstrate the superior performance of our method.

In the future, we plan to further investigate the following aspects of FSASL. 1) FSASL has three parameters to tune, which is computational cumbersome for real applications. To reduce such burden, we will replace the convex regularizations on $\mathbf{S}$ and $\mathbf{W}$ with the $\ell_0$ or $\ell_{20}$ norm. 2) FSASL is required to solve a eigen-problem, which is computational prohibitive for large scale data. Based on the connection between spectral clustering and kernel k-means \cite{ncut}, we will develop an iterative algorithm without eigen-decomposition and thus make FSASL paralleled.
\section{Acknowledgments}
This work is supported in part by the National Natural Science Foundation of China (NSFC) grants 60970045 and 60833001.

\bibliographystyle{abbrv}
\bibliography{fsasl}

\end{document}